# The Notion of Role in Conceptual Modeling

C. REYNAUD*, N. AUSSENAC-GILLES**, P. TCHOUNIKINE***, F. TRICHET***

*LRI - Univ. de Paris-Sud - Bât. 490 - 91 405 Orsay Cedex - FRANCE - cr@lri.lri.fr
** IRIT - Univ. P. Sabatier - 118, rte de Narbonne - 31062 Toulouse Cedex - FRANCE - aussenac@irit.fr
*** IRIN - Univ. de Nantes - 2, rue de la Houssinière - BP 92208 - 44322 Nantes cedex 3 - FRANCE
{Francky.Trichet, Pierre.Tchounikine}@irin.univ-nantes.fr

**Abstract:** In this article we analyse the notion of knowledge role. First of all, we present how the relationship between problem solving methods and domain models is tackled in different approaches. We concentrate on how they cope with this issue in the knowledge engineering process. Secondly, we introduce several properties which can be used to analyse, characterise and define the notion of role. We evaluate and compare the works exposed previously following these dimensions. This analysis suggests some developments to better exploit the relationship between reasoning and domain knowledge. We present them in a last section.

**Key words:** Knowledge Modeling, Knowledge Roles, Objects of Reasoning, Problem Solving Methods.

# 1. Introduction

A knowledge based system (KBS) can be viewed as a system composed of interrelated elements. The domain model describes the knowledge specific to the domain under study, the task model characterises the goals of the application and the problems to be solved and problem solving methods describe the means and control knowledge used to perform tasks. In this article we focus on the link between the domain knowledge and how it is used during problem solving (problem solving methods). This link is often ensured by roles, i.e. a description of the role that domain knowledge plays in the problem solving process [Marcus, 89]. Introduced by Mc Dermott, the notion of role appears in numerous knowledge engineering approaches, although with significantly different definitions. It may either point to the type of the domain objects that can play a role, act as a place holder for domain objects describing the role that these objects play in the problem solving process or correspond to a class which specifies an object of reasoning. These differences take their origin in the objective of the underlying modeling approach.

In this article we examine the nature of knowledge roles throughout the literature. First, we study the link between problem solving methods and domain knowledge and the benefits brought by roles during the knowledge engineering process through various representative approaches (we do not claim for an exhaustive survey): RLM, GT, KADS, OMOS, CommonKADS, TASK, EXPECT, ASTREE, VITAL-TOMAOK, MACAO-MONA, MCC and TN-PSM. Then, we introduce several properties which can be used to analyse, characterise and define the notion of role in these works. Analysing these different works in respect to these dimensions suggests developments to better exploit the relationship between problem solving and domain knowledge. We present them in a last section.

## 2. The Notion of Role

We have chronologically explored various research works, focusing on the role notion, its representation and use. Roles has been originally defined at a similar period in the role-limiting methods (RLM), Generic Tasks (GT) and KADS approaches. In more recent studies, this notion has become richer and more precise.

### 2.1. The Origins

The RLM approach has shown that acquiring knowledge to design a system is all the more efficient as the functions that knowledge plays during problem-solving are well understood [Marcus,89]. These functions, called *roles*, are specific to the problem solving method applied in the system. According to this idea, Mc Dermott's research group proposed method-oriented tools to support detailed domain knowledge acquisition. SALT, for instance, allows users to describe knowledge for design systems that apply a *propose-and-revise* strategy in terms of *parameters*, *constraints* and *revising decisions* [Marcus,89]. These *roles* represent the functional knowledge required by the *propose-and-revise* method. They specify the relevant pieces of knowledge that SALT must acquire to generate a domain-specific knowledge base. In that view, roles are method-specific but application-independent terms. In addition, they may be considered as task-independent : several tasks can be achieved by performing different methods using similar roles.

In a similar way, the GT approach emphasises on tasks and on the knowledge roles required to perform them [Bylander,88]. Roles provide a large vocabulary of task-related terms, and, additionally, tasks explicitly illustrate how domain knowledge will be used. Such a view facilitates knowledge acquisition: roles help focusing on task-relevant domain knowledge whereas the task structure guides its organisation. Note that, with GT, tasks and roles are defined in an operational shell used to design task specific knowledge bases.

The structure of *meta-class* in KADS conceptual models fulfills similar needs as roles [Wielinga,92a]. Thanks to meta-classes, the model clearly distinguishes the role of concepts during problem solving from their "meaning" in the domain. Meta-classes are defined at the inference level as the input/output parameters of inferences. They stand for domain concepts, that may play several roles, and offer a functional view on domain structures.

## 2.2. A Role as a Label

Roles have often been identified by a single term that is supposed to indicate a function in a reasoning. This trend is illustrated by the following works.

**OMOS [Linster,92]**

OMOS is a language that aims at operationalising KADS models of expertise. Building a conceptual model in OMOS includes designing a domain model (composed of classes, instances and relationships) and a method model (made of inferences and control knowledge over them). The method model is generic and domain-independent. Indeed, inferences describe actions whose input/output/control knowledge are *roles*. These two models are interrelated by specifying the mapping between domain knowledge and method roles in the inference actions. Then, solving a problem (*i.e.* running the conceptual model), dynamically builds the task model. For that purpose, the problem solver uses a *transfer* function that assigns roles to domain classes or instances, and inference actions to tuples. In OMOS, an inference action results from a two step-description:

[1] The first part defines the inference action in a generic way, by the roles it uses and the transformations it performs on domain knowledge: initialise or update a value, replace domain instances by roles or communicate with the user.

| | |
|---|---|
| (DEFINE-INFERENCE-ACTION Select-Ideal<br>WITH  INPUT-ROLE = Solution Alternative<br>       OUTPUT-ROLE = Possible-Solution<br>       CONTROL-ROLES = ((Instanciated-Criterion))<br>       VALUE-ASSIGNMENT = FALSE<br>       ROLE-ASSIGNMENT = TRANSFER | This inference requires the roles *Alternative Solution* as input, *Possible-Solution* as output and *Instanciated-Criterion* as control knowledge. TRANSFER indicates that the input domain knowledge (playing the role *Alternative Solution*) will play the role *Possible-Solution* in the following of the problem solving. |

(2) The second part describes the projection of the inference onto the domain. It refers to a particular domain relation and makes explicit the mapping between this relation arguments and knowledge roles in the inference action.

```
DOMAIN-RELATION Optimal-Clamping-Tool
WITH  ARITY = 2
      TYPE-SEQUENCE = (((INSTANCE Clamping-Tool) (EXTENSION-* Turning-Requirement)))
      ARGUMENT-SEQUENCE = (Clamping-Tool Important-Turning-Requirements))
      TUPLES = (... set of domain tuples which verify the relation)
```

This relation, defined in the domain model, has two domain concepts as parameters: an instance of *Clamping-Tool* and an Extension-* of *Turning-Requirement*. Tuples represent all the pairs of related instances according to the relation. If the inference action *Select-Ideal* uses this relation for reasoning, it must be extended as follows:

| | |
|---|---|
| DOMAIN-RELATION = Optimal-Clamping-Tool<br>   REL-ARG-TYPE =<br>       ((( 0. INSTANCE) ( 1. EXTENSION-*)))<br>   PROJ-ROLES-REL-ARGS = (((INPUT-ROLE . 0)<br>           (OUTPUT-ROLE . 0)<br>           (CONTROL-ROLES . 0) . 1))] | Argument 0 of the relation (the first one) will play input and output roles, and argument 1 will serve as control knowledge. |

To sum up, roles in OMOS are not structures. They are characterised by the way they can be assigned to domain knowledge in each inference action that uses them. This allows inferences to be described independently from the domain, and therefore enhances reusability. Furthermore, as the links between inferences and domain knowledge are made explicit, verifying the knowledge base is easier. OMOS checks that all domain concepts are associated to roles and that all domain relations required by inference actions are defined.

**CommonKADS [Wielinga,92b]**

At the inference and task levels, roles are names of object sets which share the same status in the problem solving (*design parameters, specification parameters*, etc.). Roles are used by inferences (tasks which cannot be decomposed) or transfer tasks (user interaction tasks). Three kinds of roles are distinguished. *Static roles* point to domain elements used in the problem solving process without being modified. *Dynamic roles* point towards elements that are modified during problem solving. *Case roles* are dynamic roles describing the problem to be solved. The designed model directly refers to the domain vocabulary after a mapping of (generic) terms used in the method (*roles*) onto the domain specific terms.

*Representation of roles in CML* [Schreiber,94]: CML is a structured but informal language providing a textual and graphical notation for CommonKADS expertise models. The domain knowledge description is threefold: ontologies, mapping between ontologies and domain models. An ontology reflects the structure of the domain elements at a more or less abstract level by providing several representational primitives such as concepts, relations, attributes or expressions. Domain models contain collections of domain specific statements using these primitives. In CML, roles are represented in the inference structure diagrams as the inputs or outputs of inferences or transfer tasks. In addition to their use for problem solving, they are characterised by a name and the syntactical structure of the domain items which can play them. Roles played by single domain items are distinguished from those mapped to sets of objects. Each role is locally defined in the inference where it is used.

```
inference select-parameter;
   operation-type: select;
   input-roles: parameter-set -> set of attribute-slots;
                parameter-assignments -> set of tuples
<attribute-slot, value, dependencies>;
   output-roles: parameter -> single attribute-slot;
   static-roles: formulae ∈ domain models initial-values
and calculations;
   specification: Select a parameter from the skeletal
model ………
end inference
```

In this example, parameter-set, parameter-assignments, parameter and formulae are roles which could be differently defined in another inference. On the other hand, 'attribute-slot', 'dependencies' etc. belong to the domain ontology, and 'set-of' is used to characterise the roles whose type is 'set of concepts'.

*Roles in ML²* [Van Harmelen,92] : ML² is a formal language based on logic for specifying and validating CommonKADS expertise models. In ML², a role is specified by a logical variable which is nameable (role name) and typed (sort). Indeed, ontological categories are represented by types organised in a hierarchy and defined in the domain ontology. The type characterises the structure of the associated domain items; it refers to the domain ontology (*instance, set-of-instances, etc.*). This syntactical characterisation is a classical one in logic where most of the naming operations are purely syntactical. Consequently, syntactically similar formulae have syntactically similar names. This also implies that, during problem solving at the inference level, rewriting rules allow role interpretation in the domain.

**TASK [Pierret,96]**

TASK is a coherent framework covering the whole KBS life-cycle (modeling, representation and operationalisation). The development of a KBS is viewed as an incremental process of building more and more precise and formal models. The TASK environment is composed of three languages, each of them being appropriate to one model. We only describe here tasks and methods I/O representation at the formal and operational levels. The formal specification in TASK uses Abstract Data Types (ADTs). A role is represented as a formal parameter. It can be instanciated with the application parameters if the axioms in the parameter formal specification, which represent semantic constraints, are satisfied. At the operational level, the I/O of generic tasks and methods are represented as meta-classes that implement roles. Accordingly, the I/O of the application tasks and methods are represented as classes, instances of the meta-classes defined in the generic models. Thus, in TASK, for a given application, formal specifications and operational constructs are derived by instanciating generic types or meta-classes respectively. In both cases, instanciation involves the adequacy between types or generic classes and types or application classes.

**1.3. Implicit use of the role notion**

Although they do not explicitly represent roles, some works, as the two examples below, exploit the relation between domain knowledge and how it is used during problem-solving.

**EXPECT [Swartout,96]**

As a workbench, EXPECT supports the development of KBS. In EXPECT, a method describes a particular problem-solving competence. It is characterised by its goal, its results and a body encoded with a specific language. Using reflective capabilities, EXPECT automatically analyses how methods are encoded. From the analysis of the types of the domain elements they use, it retrieves what is required to perform them. These capacities help to guide domain knowledge acquisition but also to design method-independent tools. An advantage of this approach is that the knowledge-acquisition tool is not hard-encoded. The way knowledge is used is calculated from how methods are defined, and therefore methods can be modified without modifying the tool. Moreover, new methods can be built by analogy with existing ones that share similar goals.

```
[defmethod check-capacity-constraint
    ...
    :method-body(is-smaller-or-equal
      [obj (r-capacity [r-rented-equipment ?c)))
      (than [r-volume-to-move ?c))))
```

In the body of this method, "*is-smaller-or-equal*" operates on the property *r-capacity* of an object of *r-rented-equipment* type. Then, EXPECT can detect that there is a domain object O of that type for which the method cannot be performed because *r-capacity* has no value. It can ask for this missing knowledge and justify its request by mentioning that a domain element which "plays the role r-capacity" for O is needed.

**ASTREE [Reynaud,97]**

Given an application task, ASTREE tries to automatically identify a method that achieves it. The approach assumes that the domain ontology depends a lot on the way knowledge will be used in reasoning. The idea is then to acquire first the domain ontology and to exploit it so as to find appropriate methods to achieve a target task. Even though the process it runs is generic (*select, match, ...*), the identified methods always are application-specific because derived from the interpretation of a domain ontology and from the definition of an application target task. Thus, method or task I/O refer to terms in the domain ontology. More precisely, task I/O definitions contain : [1] names of domain entities or attributes the instances or values of which respectively are the task input or output, [2] a syntactical indication about their structure (*a, a set of, a list of, sets of, lists of*), [3] constraints if syntactical characterisations of domain concepts used as input or output of the task must be specified. The constraints describe states of domain concepts with domain terms (see the example below). They can be used in different ways. They can delimit the values of attributes mentioned in the input or output part. They can also restrict the instances to those with attributes of particular values and/or to those related to some specified (or not specified) instances by a given relationship. A structured language helps to define these constraints. Thus, in ASTREE, not only the nature and structure of a concept can be specified, but its status can also be characterised. Furthermore, as ASTREE is an automatic tool, the domain ontology is represented in a formal language so that it could contain structured information useful to identify methods and the definitions of the tasks and methods are written in a structured language.

Let us illustrate the constraint language with the task of allocating rooms to employees where input data = a EMPLOYEE [e], output data = a ROOM[r]. Constraints on the output data may be imagined as follows:

"[Size-R[r] = large and Pos-R[r] = central] or Size-R[r] = small"
   if we want the allocated room to have either a large size and a central position or a small size
"contains-computer [r, Sun]" if we want to allocate a room in which a computer Sun is already set up.

## 1.4. When roles are more than labels

More recently, roles have not been considered only as labels any longer. They are more precisely characterised by additional properties, as in the following approaches.

### VITAL - TOMAOK [Leroux,95]

As part of the VITAL project, TOMAOK developed a rewrite grammar to refine an application problem-solving model. The approach is based on a tight coupling between domain and problem solving. To design a model, generic inference primitives coming from generalised directive models (GDM) are reused and adapted. Each model component progressively evolves. To refine the inference structure and the meta-view of the domain, the rules examine the roles or the domain element types which the roles stand for. This may lead to introduce new inferences and new roles, or to characterise better roles or domain elements. The notion of role is essential in TOMAOK. Domain elements have to satisfy the syntactical and semantic constraints required by roles. Moreover, role properties specify the conditions in which a generic method can be applied and decomposed to model a given application.

In GDM, the role description renders, at a meta-level, the domain knowledge organisation. It reflects the ontological commitments implied by the other categories of the model. All the descriptions form what is called the meta-domain, a component that does not exist in KADS expertise model. A role is a strongly typed structure characterised by several types: one characterises its use in the inference structure (static, dynamic, case, dynamic-control, case-control) ; a syntactical one defines constraints on domain elements (their type and their syntactical structure) ; a semantic one can be either a role of the method ontology (i.e. *class of decision, finding*) for instances, values or concepts, either "logical" or "calculated" for expressions, or the name of a relation for structures. Roles are represented with frames in the I/O part of each inference. Beside its types, each role contains the definitions of all the concepts which it can map to. These definitions are specific to each role ; each concept only contains the attributes required in the inference where the role is defined.

For example, the role SOLUTION is defined in several inferences: it is generated by INSTANCIATE and is an input of VERIFY. SOLUTION is defined the same way in all these inferences. However the definitions of the associated concepts, which may appear in other roles, only mention the properties which are useful in the inference where the definition is given.

Role Name: SOLUTION
Role Type: Dynamic Role
Denoted Knowledge Syntactic Type: Set of Concepts
Denoted Knowledge Semantic Type: Correspondence
Denoted Knowledge Qualification:
Role Domain Definition:
    Concept Name: ROOM
    Concept Properties:
        Property Name: NAME

Property Type: IDENTIFIER
  Type definition: String - Values Range:
Property Name: SIZE
Property Type: ROOM SIZE

Type definition: Enumerated - Values Range:
[Single Twin] ....
Concept Name: EMPLOYEE
Concept Properties: ...

### 3.1. The Role Notion in MACAO [Aussenac,94a], [Aussenac,94b]

MACAO is a general methodology for knowledge acquisition and modeling which provides techniques and guidelines to knowledge engineers. Applying MACAO results in designing a two-fold conceptual model, where domain knowledge (a conceptual network) is separated from problem solving knowledge (a task-method decomposition tree).

**Roles in MONA [Matta,95]**

Knowledge is represented using MONA, a conceptual formalism specific to MACAO and adapted from the LISA operational language [Delouis,95]. The MACAO workbench offers tools for detailed and schematic knowledge elicitation as well as knowledge and diagram editors to structure knowledge in a model and represent it with MONA. The MONA primitives are quite basic in conceptual modeling: concepts, typed relations and expressions for domain modelling ; tasks, methods and roles to represent problem solving knowledge. Each primitive is represented with a frame, described both in natural language and by its relations with other structures. To this extent, MONA is similar to CML.

Roles have been defined to represent the method and task inputs and outputs with a domain independent vocabulary. This structure has been introduced to make possible connections between domain and problem solving knowledge explicit. By this means, generic problem solving methods can be refined to design an application problem solving model. Further-more, the problem solving model is more efficient to guide knowledge elicitation. Roles are characterised by their name and the set of domain items that can play them: concepts, concept attributes or their values. However, because MONA is a conceptual and not an operational language, the mapping of roles to domain items was not completely specified. This problem arose during the ZTM experimentation [Beaubeau,96].

**An operationalisation of roles: ZTM**

The goal of the ZTM project was to define, with the ZOLA operational language [Trichet, 97], operational structures corresponding to MONA primitives. However, running the model required to better define roles. In addition to their global definition, roles should convey information about which of the possible domain items is actually playing it at a step of the problem solving. So it seemed necessary to keep track of the context in which domain knowledge could play a given role, by indicating the input or output of the task or method where this role was used. Consequently, in ZTM [Beaubeau,96], roles are now represented by specific objects with a name, the task or method characteristics where they are used, and, for each, the different possible domain values that can play this role. This value can be a concept or a role, depending on whether the role is played by domain knowledge or by another role.

Role Name: set-of-components
Possible Values:
      Value 1   Link to a domain concept: set-of-persons
                    Related characteristic: input-context, task Assign-places-to-components
      Value 2   Link to a role: [set-of-components, input-context, task Assign-places-to-components]
                    Related characteristic: parameter, method Assignment-with-criteria
      Value ...

In this example, *set-of-components* describes, at an abstract level, the input of task *Assign-places-to-components* [value 1] and the parameters of method *Assignment-with-criteria* [value 2]. In value 1, *set-of-components* can be assimilated to a static role which points to d*set-of-persons* when used as the input of task *Assign-places-to-components*. In value 2, *set-of-components* can be considered as a dynamic role ; there is no explicit link to domain knowledge, but a reference to another role : the parameters of method *Assignment-with-criteria* are the *set-of components* used as input of task *Assign-places-to-components*.

### MCC [Causse,93]

In the MCC approach, K. Causse proposes to add a level for the description of roles, which are considered as reasoning entities manipulated by the reasoning process. This level is the heuristic level. It is at the interface of the domain level and the reasoning level. This heuristic level allows the natures of the roles to be specified and meta-information on the roles to be expressed as attributes of the reasoning entities. This way, roles are neither embedded inside the methods of problem solving nor assimilated to the domain knowledge. They have a specific status and can be studied as such. We present below an illustration of a MCC structure equivalent to the notion of role : a schema.

**schema** hypothesis
    **com:** an hypothesis holds for a system state which may cause the observed symptoms;
    **refers to:** state;
    **rel:** cause-p: hypothesis or complaint;
    **rel:** inv-cause-p: hypothesis;
    **rel:** stated-by: evidence;

In this example, the "hypothesis" structure is domain-independent, except its attribute "refers to". Indeed, usually, schemas are expressed in reference to problem solving methods. This "hypothesis" structure has been defined for the systematic diagnostic method but not in reference to the domain of car diagnosis, which was the one of the application.

### The TN-PSM Approach [Beys,96]

Recently, so as to facilitate reuse of similar problem solving methods across different tasks, Beys proposes to represent methods using task independent terms. These methods are called Task Neutral Problem Solving Methods (TN-PSM). Each TN-PSM contains assumptions which specify syntactical constraints over the structure or the properties of the manipulated knowledge as illustrated below. Pursuing this approach, [Beys,96] suggests that the usability versus reusability trade-off can be tackled by automating the process of uncovering the domain structure required by the method.

**psm** cover;
    **input:**        leaves: Sets of leaf nodes that need to be covered;
    **output:**      cover: Sets of root nodes that cover the leaves;
    **competence:** ...
    **sub-tasks:** ...

```
    additional roles: ...
    control-structure: ...
    structure-assumptions:   ∃ X in the ontology such that: X: sequence($x_1, x_2, ..., x_n$),
                             ∀i ∈ [1,n] binary-relation ($x_i, A_{i-1}, A_i$)
                             ∀ i ∈ [1, n-1], class($A_i$) $A_0$ = set of leaves, $A_n$ = set of roots
    property-assumptions: ∀ i ∈ [1,n] if $A_{i-1}$ = $A_i$, asymmetric ($x_i$), irreflexive ($x_i$), transitive ($x_i$) endif
                          one-to-many relation (X);
end psm cover;
```

## 2. Analysis of the Notion of Role

We have identified several criteteria to analyse and characterise roles in the previous works. In the following, we present the results of our analysis.

### 2.1. The dimensions of our analysis

**The Modeling Process**

The representation and the use of roles differ according to the knowledge modeling process. A priori, knowledge roles are not useful when modeling consists in designing a conceptual model from the domain interpretation such as in ASTREE. Nevertheless, this notion can be introduced to verify the model, to support the acquisition of detailed knowledge (MACAO) or to acquire and explicitly represent control knowledge (MCC). On the other hand, the notion of role is essential when a top-down modeling approach (partial or not) is adopted, i.e. when global generic models are reused (GT) or different components are reused (CML, RML), combined and refined (TOMAOK). In such approaches, the primitives of the generic components (and mainly the roles) are described in an abstract way so as to be applicable across various domains (GT, RLM, CML) or even various tasks (TN-PSM).

**The Definition, Description and Localisation of a Role**

The early approaches to knowledge level modeling (RLM, GT, KADS) focus on model-based reasoning. They emphasise the distinction between the knowledge level and the implementation level. The problem-solving process is described at an abstract level. The required knowledge is denoted by its role in reasoning (a brief commentary gives the meaning of each role), providing guidance in knowledge acquisition (what knowledge should be acquired and which it will play) and enabling reuse of generic constructs. In such a context, domain models are not emphasised. Thus, the choice of precise and proper names for roles, denoting precisely their place in reasoning, is essential. In these approaches the connection between domain and problem solving is defined as a projection (RLM), a mapping relation (OMOS) or an instanciation process (TASK). This way, each role is a functional label which is linked to a set of domain concepts.

Later, the interest of combining the reuse of generic components and data abstraction has been recognised. A stronger coupling between domain and problem solving is then necessary. As a consequence, the semantics of the roles must be described more precisely. This leads, for example, to add indications about the semantic type of the knowledge required by a method to play a role (TOMAOK). This also leads to the characterisation of the roles with particular attributes (MCC).

More recently, some works proposed to add syntactical properties or to substitute the semantic definition of a role by a syntactical one (TN-PSM, MCC). These properties can refer to the structure of the required knowledge or to mathematical properties (reflexive, symmetric, transitive, etc.). Syntactical properties of a role allow to partially automate mapping between domain knowledge and problem-solving. Indeed, various elements of domain knowledge can satisfy the same syntactical properties while all these elements may not be equally adequate to play a given role. In that case, when the semantics of a role is totally ignored, a domain expert has to interpret the results of the mapping process.

Finally, whatever the description of a role is, its representation can be either local (OMOS, TOMAOK) to a task or a method or global to the whole model (MCC). In most approaches it is local but sometimes, roles are viewed as independent reasoning objects. Furthermore, the connection between domain and problem solving is often a link between roles and concepts defined at the domain level. However, definitions of domain concepts can also be included in task or method definitions (OMOS, TOMAOK).

**The Level of Formalisation**

Whatever the nature of the role definition, whether it is represented with an indenpendant primitive or not, it can be represented using a formal language or not. As a consequence, the range of formalisation levels goes from very weak formalisms such as frames woth natural language descriptions (CML, MACAO) or frames with a formal language (TOMAOK, MCC) to operational primitives (OMOS, RLM, GT) or logical predicates (ML$^2$, TS-PSM).

This dimension is crucial. However, we do not intend to report here the debate concerning the interests and limitations of a formal model versus an unformal one. The same arguments apply for roles, which are part of the modeling language : unformal representations are obviously relevant in the early steps of knowledge acquisition and modeling, whereas formal representations provide more guidance either in the testing and validation steps, or to acquire detailed knowledge and refine a model.

## 2.2. The use of Roles in the Modeling Process

It is interesting to identify the various reasons that motivated the introduction of the role notion in each modeling approach. We have listed the main kinds of guidance that role may provide in a knowledge engineering environment :

- to acquire additional knowledge: acquire detailed domain knowledge that instanciate the problem-solving method (RLM, GT) or refine the problem-solving method itself (OMOS);

- to test and check the model: either to check if all the domain knowledge required by the problem-solving method is available (RLM, EXPECT) or to validate the model structure by running it (OMOS, ML$^2$) or by showing an abstract and easy-to-understand representation of the problem-solving process (MACAO, CML);

- to reuse generic components (GT, KADS, RLM, TASK) ou to make it easier to reuse specific components (MACAO, MCC).

## 3. Discussion

Our analysis has shown that the definition and description of roles closely depend on why they have been defined in the modeling approach. Given the guidance that we expect from a modeling environment, we promote a double characterisation of roles.

### 3.1. Connection Between how Roles are Represented and their use

A role can be a means to guide the acquisition of different kinds of knowledge, in particular when the modeling approaches are model-driven. The problem solving method is then viewed as a skeletal model, in which generic knowledge requirements are defined, and that has to be instanciated. The expert is asked about domain concepts, instances or control knowledge that play roles, according to their meaning denoted by their names or to their definition. Thus, roles provide strong guidance for the model instantiation process. They are the means by which knowledge acquisition tools can interact with the user in method-oriented terminology. Such tools prompt the user for method-specific knowledge types and can prevent him from entering expressions that conflict with the definitions of these types in the problem-solving method. These tools are all the more efficient as roles are well structured. Automatic matching with domain knowledge according to syntactical or semantical criteria may also be possible with a formal representation of roles. However, a good semantical characterisation is difficult to achieve.

When the modeling process enables a distinct representation of various types of knowledge (*i.e.* domain, task and inference knowledge), roles can be suitable for verification (OMOS, GT, RLM). All domain concepts must be associated to roles, all the relations required by tasks must be defined at the domain layer. In the particular case where the notion of role is not explicit (EXPECT, ASTREE), though manipulated by reflective tools, methods directly refer to domain concepts. Then we can verify that their instances are defined in the domain model and that required attributes are valued. Moreover, other kinds of verifications are possible when using formal specifications (ML$^2$, TASK). They help check the correspondance between the types of roles and the types of the domain concepts they stand for. Formal specifications can also be used to prove the correctness of a model as done in software engineering. Finally, the operational description allows prototyping to evaluate the behaviour of the KBS under development (ZTM, OMOS).

Roles can also be a means to support the conceptual model design. The overview shows two illustrations. First of all, roles can help to select a way to expand an initial abstract problem-solving model (TOMAOK, OMOS). Thus, they provide guidance to progressively refine a selected model that defines precise knowledge requirements or knowledge roles. Roles specify the syntactical and semantical constraints that associated domain concepts must satisfy. Properties abd types of the elicited knowledge are criteria to select the applicable rule to refine the initial model. Secondly, roles can facilitate the modeling process thanks to the representation of control knowledge (MCC). In that case, roles are reasoning entities defined as such with attributes expressing meta-information. They point to domain knowledge but additional knowledge that would not exist outside the reasoning scope can also be represented. This representation allows control knowledge to be clearly defined and labelled. It is a way to facilitate the construction of problem solving methods.

Reuse being a current important issue in knowledge engineering, many works aim at enabling partial reuse of problem-solving models. The implications of this objective on roles differ according to whether the reuse scope is a task-specific context (GT, KADS) or not (TN-PSM). The inference primitives can be described at two levels: a generic one, which qualifies the I/O according to their function in the inference and in the task to be achieved, and another one which establishes a link with domain concepts. On the contrary, task-neutral approaches propose to describe roles in a merely syntactical manner: their name is very general and they are not linked to domain concepts. Such a representation is supposed to make reuse of similar methods across different tasks easier. Reuse is also an important dimension for role because of its impact on the representation language. Formal specifications facilitate a clear understanding of roles. They can be useful for specifying reusable methods. Finally, reuse also depends on the description of roles. A role is all the more reused since it is precisely defined, both with syntactical and semantical properties.

## 3.2. Towards a double characterisation of roles

The advantages provided by the numerous uses of the roles convinced us to consider this notion in a problem-solving model (PSM). Roles are different from domain concepts and must be clearly distinguished because they are of another epistemological type. They make sense only in the context of PSM, not in the context of application models. On the opposite domain concepts are not dependent of any reasoning. In addition, we consider that roles have to be represented as specific structures. That way, they can be more easily manipulated by the reasoning process and easily characterised with intrinsic properties.

Furthermore we consider that roles must be characterized both from a syntactical and a semantical point of view. Their syntactical characterisation may correspond to structural or property assumptions (TN-PSM). Yet, unlike the current advances that promote the syntactical characterisation of roles against their semantic labelling, we consider that their task-oriented name and semantic characterisation should not desappear. When knowledge modelling combines abstraction and reuse, the two aspects are important. In a bottom-up context, the model is interpreted by both the expert and the designer. The semantics of the roles plays a major part. Role labels would be all the more understandable as they are defined in keeping with the *task* problem-solving model. It will be also easier for the expert to guide the mapping between domain knowledge and roles. In the TN-PSM approach, the domain knowledge is automatically mapped to the methods by testing the syntactical constraints. Thereafter, a domain expert has to interpret the results of the mapping. But how can it be done without precising the semantics of the roles ?

This conclusion leads us to discuss the ways the semantics of roles can be defined. In many research works, this semantics is only denoted by a name with a brief comment on the role meaning. So the semantics has to be intuitively deduced from the name. To make this process more powerful, the role names sometimes refer to terms in method ontologies which describe them (TOMAOK). This assumes that such ontologies are available. In MCC, roles have their own attributes, which contributes to define their semantics. We could also imagine to build state models (SM) to define them. In most object-oriented analysis methodologies, SMs specify the various states of the processed objects as well as the events that make an object evolve from one state to another. In a similar way, the semantics of a role could be defined by all its possible states at the different times of the reasoning process and by the methods or inferences which make it change from one state to another. Then, a SM would be specific to a given PSM. This view is borrowed from ASTREE in which the notion of state is defined for the domain concepts in the I/O of tasks. That way, a role could be identified by the name of a reasoning object in a particular state.

Finally, the definitions of the roles must be supplemented with indications on the types of the domain concepts which can play them. Very often, the connection between domain and problem solving is a mapping relation between roles and domain concepts. Our analysis convinced us that this mapping must be more precise. It is in fact a relation between roles and domain concepts *in a certain state*   (ASTREE).

## 4. Conclusion

In this article, we have stressed the key importance of roles at the frontier between domain and problem solving knowledge in conceptual modeling. They help make explicit some of the implicit assumptions, also called ontological commitments, that problem solving methods impose to domain knowledge. Several uses of the notion of roles can be identified. We have shown that they affect its existence, its characterisation and its representation.

An exploration of the state of the art has revealed a tendency that promotes a formal syntactical description of roles in order to broaden the reusability of the methods that use them. Moreover, these works consider roles as a sub-product of inference actions. Our position is quite different. First of all, we consider that roles should be isolated as such, in specific structures, so that they could be easily identified and that a system could reason about them. Secondly, we claim the advantages of keeping a semantics in the roles to facilitate their interpretation and to increase their interest in non-generic models. This is why we have investigated possible ways to define this semantics. Thirdly, and for this purpose, we consider that the state of roles is crucial. Furthermore, the states of the domain concepts are to be considered in the mapping relation between roles and domain concepts.

However, practically implementing such a notion raises several major issues, still at the heart of the debate [Van Heijst,97] that we have not answered yet: how to decide of a role label ? role definitions might be redundant with concept specification: how can we avoid it ? It increases the number of structures in the conceptual model: is it realistic ? We feel like experimenting our approaches with the new definitions we gave to roles. By this means, we will test our hypotheses and certainly have some answers to these questions.

## References

Aussenac-Gilles N., "How to combine data abstraction and model refinement: a methodological contribution in Macao". *A future for Knowledge Acquisition*, Proc. of EKAW94. Berlin: Springer Verlag. LNAI 867, pp. 262-282. 1994.

Aussenac-Gilles N., Matta N., " Making the method of problem solving explicit with Macao: the Sisyphus case-study ". *International Journal of Human Computer Studies*. 40, pp. 193-219. 1994.

Beaubeau D., Aussenac N., Tchounikine P., "Mona au pays des roles". Rapport IRIT/96-23-R, Juil. 96.

Beys B., Benjamins V. R., Van Heijst, G., " Remedying the Reusability - Usability Trade-off for Problem Solving Methods". *Proc. of KAW'96*, Gaines & Musen ed., Banff: SRDG Publications Univ., Nov. 1996.

Bylander T., Chandrasekaran "Generic tasks in knowledge-based reasoning: the right level of abstraction for knowledge acquisition". In B.R. Gaines & J. H. Boose, Eds. *Knowledge Acquisition for Knowledge-Based Systems*. Vol. 1, pp. 65-87. London: Academic Press. 1988.

Causse K., " Heuristic control knowledge ". *Knowledge Acquisition for Knowledge Based Systems*. Proc. of EKAW'93. Aussenac N., Boy G., Gaines B., Linster M., Ganascia J.G., Kodratoff Y. Eds. LNAI 723. Heidelberg: Springer Verlag. pp. 183-199. 1993.

Delouis I. Krivine J.P., " LISA, un langage réflexif pour opérationaliser les modèles d'expertise ". *Revue d'Intelligence Artificielle*. Volume 9:1. pp. 53-88. Paris: Hermès. 1995.

Istenes Z., Tchounikine P., " Zola: a language to Operationalise Conceptual Models of Reasoning". *Journal of Computing and Information* [Proc. of ICCI'96] 2:1, pp. 689-706, 1996.

Leroux B., Laublet P., " Steps towards a Unified Approach to Knowledge Modelling". *Proc. of VIth European Japanese Conf. on Information Modelling and Knowledge Bases.* Amsterdam: IOS Press. 1995.

Linster M., " Knowledge Acquisition Based on Explicit Methods of Problem Solving ". PhD Thesis, Univ. of Kaiserslautern. Feb. 1992.

Marcus S. McDermott J., " SALT: a knowledge acquisition tool for propose and revise systems". *Artificial Intelligence*, 39, pp. 1-37, 1989.

Matta N., " Méthodes de résolution de problèmes: leur explicitation et leur représentation dans Macao-II". Thèse de l'Université P. Sabatier, Toulouse. Oct. 1995.

Pierret-Golbreich C. Talon X., " TFL, an algebraic language to specify the dynamic behavior of Knowledge-Based Systems ". *The Knowledge Engineering Review*, Volume 11:3. pp. 253-280. 1996.

Reynaud C., Tort F., " Using Explicit Ontologies to Create Problem Solving Methods ". *International Journal of Human Computer Studies*, 46. pp. 339-364. 1997.

Schreiber G., Wielinga B., Akkermans H., Van de Welde W., Anjewierden A., "CML: The CommonKADS Conceptual Modeling Language ". *A future for Knowledge Acquisition*. Steels L., Schreiber G., Van de Velde W., eds. LNAI n°867. Berlin: Springer Verlag. 1994. pp 1-25.

Swartout B., Gil Y., "Flexible knowledge acquisition through explicit representation of knowledge roles". AAAI Spring Symp. Acquisition, learning and demonstration: automating tasks for users. Stanford (USA). March 1996.

Trichet F., Tchounikine P., " Reusing a Flexible Task-Method Framework to Prototype a Knowledge Based System ". Proc. of *Software Engineering and Knowledge engineering* (SEKE'97). Madrid, june 1997.

Van Harmelen F. Balder J., "(ML)$^2$: A Formal language for KADS models of expertise ". *Knowledge Acquisition*, 4 (1), pp. 127-161, 1992. Special issue : 'The KADS approach to knowledge engineering'

Van Heijst G., Schreiber A., Wielinga B. " Roles are not classes: a reply to Nicola Guarino ". *International Journal of Human Computer Studies*. 46, pp 311-318. March 1997.

Wielinga B., Schreiber G., Breuker J., " KADS: a Modeling Approach to Knowledge Engineering". *Knowledge Acquisition*, 4(1). pp. 5-54, 1992. 'The KADS approach to knowledge Engineering".

Wielinga B.J. Van De Velde W. Schreiber G. Akkermans H., "Towards a unification of knowledge modelling approaches". In David, J.-M., Krivine, J.-P., Simmons R., ed., *Second Generation Expert Systems*, Springer-Verlag. 1992.